\documentclass{article}

\usepackage{microtype}
\usepackage{graphicx}
\usepackage{subfigure}
\usepackage{booktabs} 

\usepackage{hyperref}

\usepackage[accepted]{icml2025}

\usepackage{amsmath}
\usepackage{amssymb}
\usepackage{mathtools}
\usepackage{amsthm}

\usepackage[capitalize,noabbrev]{cleveref}

\theoremstyle{plain}

\theoremstyle{definition}

\theoremstyle{remark}

\usepackage[textsize=tiny]{todonotes}

\icmltitlerunning{Fine-Tuning Next-Scale Visual Autoregressive Models with Group Relative Policy Optimization}

\begin{document}

\onecolumn

\icmltitle{Fine-Tuning Next-Scale Visual Autoregressive Models \\ with Group Relative Policy Optimization}




\begin{icmlauthorlist}
\icmlauthor{Matteo Gallici}{upc}
\icmlauthor{Haitz Sáez de Ocáriz Borde}{Cambdrige}
\end{icmlauthorlist}

\icmlaffiliation{upc}{Universitat Politècnica de Catalunya}
\icmlaffiliation{Cambdrige}{University of Cambdrige}


\icmlcorrespondingauthor{Matteo Gallici}{gallici@cs.upc.edu}

\icmlkeywords{Reinforcement Learning, Image Generation, Autoregressive Model, Fine-Tuning, Machine Learning, ICML}

\vskip 0.3in



\printAffiliationsAndNotice{}  

\vspace{-30pt}
\begin{figure*}[ht!]
    \centering
    \includegraphics[width=0.99\textwidth]{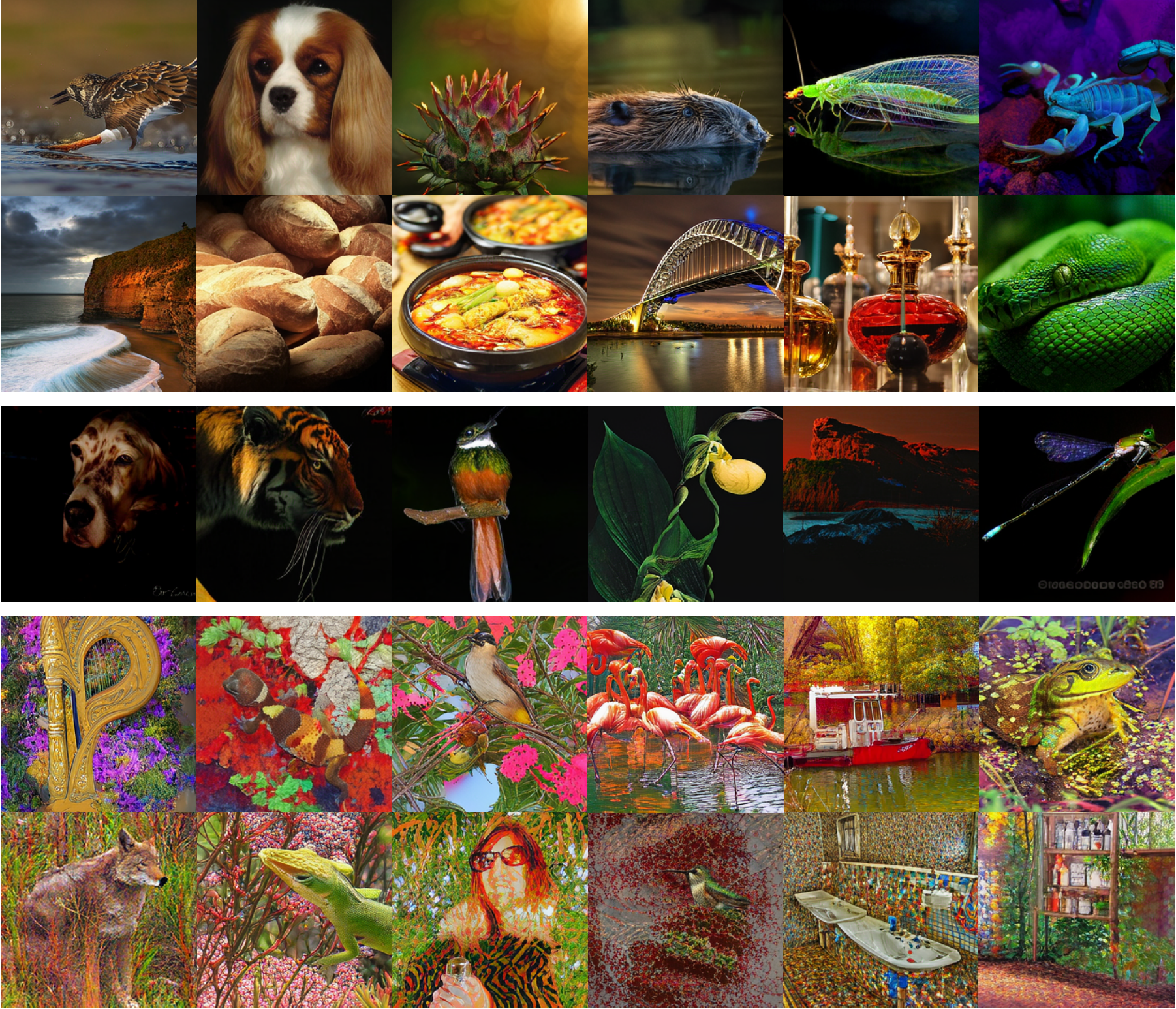}
    \caption{Images from VAR-\textit{d}30~\citep{VAR} fine-tuned with group relative policy optimization. At the top, samples from the model fine-tuned with an aesthetic reward. In the middle, samples from the model fine-tuned to produce paintings optimized with CLIP score. At the bottom, the model is fine-tuned with a combination of the two rewards. Given the notable scarcity of artworks in the ImageNet dataset used for pre-training VAR, the artistic patterns observed in the images should primarily be attributed to discoveries made during fine-tuning through reinforcement learning.}
    \label{fig:banner_image}
\end{figure*}

\vspace{2pt}

\clearpage

\twocolumn

\begin{abstract}
Fine-tuning pre-trained generative models with Reinforcement Learning (RL) has emerged as an effective approach for aligning outputs more closely with nuanced human preferences. In this paper, we investigate the application of Group Relative Policy Optimization (GRPO) to fine-tune next-scale visual autoregressive (VAR) models. Our empirical results demonstrate that this approach enables alignment to intricate reward signals derived from aesthetic predictors and CLIP embeddings, significantly enhancing image quality and enabling precise control over the generation style. Interestingly, by leveraging CLIP, our method can help VAR models generalize beyond their initial ImageNet distribution: through RL-driven exploration, these models can generate images aligned with prompts referencing image styles that were absent during pre-training. In summary, we show that RL-based fine-tuning is both efficient and effective for VAR models, benefiting particularly from their fast inference speeds, which are advantageous for online sampling—an aspect that poses significant challenges for diffusion-based alternatives.
\end{abstract}

\section{Introduction}

For Artificial Intelligence (AI) systems to integrate successfully into real-world applications and products, it is crucial to design intelligent systems that support effective user interaction, align with human preferences, and exhibit ethical, user-centric behaviors. This alignment is particularly challenging in generative modeling; although recent breakthroughs have dramatically advanced image synthesis, ensuring generated images adhere to nuanced human aesthetic and semantic criteria remains difficult. In this context, Reinforcement Learning (RL)-based fine-tuning has emerged as a powerful tool for aligning pre-trained generative models with human preference, gaining prominence first in language~\citep{ouyang2022traininglanguagemodelsfollow}, and more recently in the image domain~\citep{yang2024usinghumanfeedbackfinetune,black2024trainingdiffusionmodelsreinforcement}.

Current methods predominantly apply RL to diffusion-based models due to their state-of-the-art generation quality. However, these models typically incur substantial computational costs during sampling, significantly hampering their efficiency for online RL fine-tuning. Autoregressive models, particularly those leveraging a next-scale approach such as VAR~\citep{VAR}, provide a promising alternative by enabling much faster inference while maintaining competitive image quality. Motivated by the advantages of autoregressive modeling, we explore Group Relative Policy Optimization (GRPO)~\cite{shao2024deepseekmathpushinglimitsmathematical}, a newly introduced RL variant based on Proximal Policy Optimization~(PPO)~\citep{schulman2017proximalpolicyoptimizationalgorithms}, to fine-tune VAR. GRPO eliminates the need for a separate value function estimator through grouped sampling, enhancing both efficiency and training stability. Concretely, our contributions are the following:
\begin{itemize}
\item We propose the \textit{first framework integrating GRPO with next-scale visual autoregressive models (VAR)}, leveraging the computational efficiency and sampling speed of VAR.
\item We empirically demonstrate that VAR models fine-tuned with GRPO significantly \textit{improve image aesthetics and alignment} with complex reward signals derived from CLIP and human-rated aesthetic preferences.
\item We provide evidence that GRPO fine-tuning enables VAR models to \textit{generalize beyond the original ImageNet training distribution}, effectively aligning to prompts referencing unseen concepts and aesthetic criteria through RL-driven exploration.
\end{itemize}

\section{Related Work}

Before presenting our approach, we first discuss related work in image generative modeling and RL-based alignment of generative models, both in the context of language and image modeling.

\subsection{Image Generative Models} 

Autoregressive~\citep{Oord2017NeuralDR}, diffusion~\citep{SohlDickstein2015DeepUL}, and flow matching models~\citep{lipman2023flow}, distinguish themselves from other generative modeling approaches, such as Variational Autoencoders~(VAEs)~\citep{Kingma2013AutoEncodingVB} and Generative Adversarial Networks~(GANs)~\citep{goodfellow2014generativeadversarialnetworks}, by simplifying the task of generating samples from complex data distributions in a single forward pass into a series of smaller, more tractable intermediate predictions. 

\paragraph{Diffusion and Flow Matching} Diffusion~\citep{SohlDickstein2015DeepUL} and flow matching~\citep{lipman2023flow} (or rectified flows~\citep{liu2023flow}) have been highly successful in image generation using large scale models~\citep{rombach2022highresolutionimagesynthesislatent,podell2023sdxlimprovinglatentdiffusion,esser2024scaling}. However, they require generating intermediate trajectories between the base and target distributions with the same dimensionality as their final output, which can become increasingly computationally expensive as the dimensionality of the data they are training on grows (even when operating in the latent space of an autoencoder). This can also make RL fine-tuning challenging due to slow sampling, as we will discuss later.

\paragraph{Autoregressive Image Models} Autoregressive modeling~\citep{Oord2017NeuralDR}, on the other hand, relies on a language-modeling-style next-token prediction approach, which can make inference substantially faster. Autoregression achieves this by representing the data as a token sequence and predicting each element successively. For instance, the original DALL-E model~\citep{ramesh2021zeroshottexttoimagegeneration} was based on such an approach. In particular, a next-scale based variant of the original autoregressive paradigm has gained recent attention in the literature with the introduction of class-conditioned image models such a VAR~\citep{VAR}, as well as text-to-image models like Infinity~\citep{Infinity}, and Switti~\citep{voronov2024switti}, to name a few. This will be the focus of our paper.

\subsection{Generative Model Alignment with Reinforcement Learning}

Using RL to steer the behavior of large pre-trained models has proven very successful in language and has only recently gained traction in other data modalities.

\paragraph{Large Language Model Alignment} Defining what a human considers to be a `good' text sample can be very challenging to do via a loss function. Instead, LLMs are first pre-trained on a vast corpus of internet-scraped text data, minimizing a cross-entropy loss to acquire their base knowledge, and then later aligned through fine-tuning on instructions. Since the introduction of Reinforcement Learning from Human Feedback~(RLHF)~\citep{ouyang2022traininglanguagemodelsfollow} to turn base language models into chatbots (e.g. GPT-3~\citep{brown2020languagemodelsfewshotlearners} into ChatGPT), there has been a plethora of follow-up research that has ultimately lead to using RL to teach LLMs how to reason. RLHF leverages a reward model, that is, a neural network that takes in text tokens and returns a scalar value that numerically represents human preference. In this setup, the policy is the language model itself, generating text sequences based on a given prompt. Its action space is the token vocabulary size, and its observation space is all possible input token sequences. Finally, the reward function guides the learning process, combining feedback from the aforementioned reward model with a mechanism to prevent drastic changes to the policy, often based on the Kullback–Leibler (KL) divergence. We will be working with a similar setup in Section~\ref{sec:Method}.

\paragraph{Image Model Alignment} Aligning image diffusion models with similar techniques presents a challenge: their intractable exact likelihood computation makes it difficult to directly apply many conventional RL algorithms. While recent literature has proposed diffusion-specific frameworks—such as reward-weighted likelihood maximization~\citep{lee2023aligningtexttoimagemodelsusing}, Direct Preference for Denoising Diffusion Policy Optimization (D3PO)~\citep{yang2024usinghumanfeedbackfinetune}, Denoising Diffusion Policy Optimization (DDPO)~\citep{black2024trainingdiffusionmodelsreinforcement}, and very recently, GRPO-based approaches like Flow-GRPO~\citep{liu2023flow} and Dance-GRPO~\citep{xue2025dancegrpounleashinggrpovisual}—we argue that this overcomplication would be removed if the image generative model were autoregressive. Additionally, the latter significantly improves the computational amenability of RL fine-tuning due to its fast sampling speed. Although recently \citep{wang2025simplear} explored the use of GRPO in a simple autoregressive multi-modal model, the application of RL in next-scale autoregressive prediction remains unexplored. 

\section{Preliminaries}
\label{sec:Preliminaries}

In this section, we review the next-token prediction paradigm and the next-scale formulation for image generation.

\paragraph*{Next-token Prediction} 
We use the term \textit{token} to denote the index of an entry in the codebook (or vocabulary used by our model). Consider a sequence of tokens $[e_1, e_2, \dots, e_T]$, where $T$ denotes the sequence length. The next-token (causal) autoregressive model formulation assumes that the probability of a token $e_t$ is conditioned solely on the preceding tokens $[e_1, e_2, \dots, e_{t-1}]$, establishing a one-way dependency among them. This allows the likelihood of the entire sequence to be factorized as: $p([e_1, e_2, \dots, e_T]) = \prod_{t=1}^{T} p(e_t \mid [e_1, e_2, \dots, e_{t-1}]).$ Autoregressive models learn to model data by maximizing the likelihood of observed sequences during training, $p_{\theta}([e_1, e_2, \dots, e_T])$. Specifically during large-scale pre-training, the model's parameters $\theta$ are adjusted to minimize the negative log-likelihood, typically through the use of cross-entropy loss: $\mathcal{L}(\theta) = -\sum_{i=1}^N \log p_{\theta}([e_1, e_2, \dots, e_T]_i),$ where $[e_1, e_2, \dots, e_T]_i$ refers to the $i$-th training sequence out of a total of $N$. By minimizing $\mathcal{L}(\theta)$, the model learns to predict each token in the sequence based on its preceding context, a process commonly known as \textit{next token prediction}. After training, the model can generate new sequences based on the learned probability distribution. This formulation is especially natural in language modeling, where the text exhibits a clear causal structure.

\paragraph{Visual Autoregressive Modeling} Traditionally, the next-token prediction paradigm has been adapted for image generation by flattening visual samples, predominantly using a raster-scan approach, where image patches are arranged into a sequence~\citep{Oord2017NeuralDR}. This disrupts the local spatial correlations in images and other visual data, such as videos, hinders the model's ability to learn bidirectional correlations within the sequence, and may also negatively affect zero-shot generalization. For instance, it restricts lower-right image patches to depend on upper-left patches but not vice versa. Visual AutoRegressive modeling (VAR)~\citep{VAR} suggests using a more natural multi-scale, coarse-to-fine representation for images. In VAR the likelihood of the data sample is modeled as follows: $p([r_1, r_2, \dots, r_K]) = \prod_{k=1}^{K} p(r_k \mid [r_1, r_2, \dots, r_{k-1}])$, where $r_k$ is used to denote an autoregressive unit at each scale $k$ of the visual data representation consisting of $h_k \times w_k$ tokens, where $h_k,w_k \in \mathbb{N}_{+}$ and $h_K>h_{K-1}>...>h_1$ and $w_K>w_{K-1}>...>w_1$. In short, in this formulation we impose a causal autoregressive dependency at the resolution (or scale) level, rather than at the token level. That is, $r_k$ is a collection of tokens instead of a single token. This, in turn, requires to construct an alternative latent space, where the continuous embedding is represented using quantized multi-scale residual embeddings. Similar to next-token prediction, once the transition probabilities have been learned, it can be used to generate new data samples via the token-to-codebook mappings and passing the final embedding reconstruction to a decoder.


\section{Method}
\label{sec:Method}

As mentioned, we specifically focus on applying RL to next-scale autoregressive image models. Inspired by successful RL applications in language modeling, we observe a natural alignment since both language models and autoregressive image generators follow a next-token prediction paradigm. Additionally, the superior efficiency and speed of autoregressive models make them particularly well-suited to online RL, where online sampling typically presents the main computational challenge.

\subsection{Group Relative Policy Optimization (GRPO)}
\label{subsec:Group Relative Policy Optimization (GRPO)}

Among the techniques that were recently proposed to fine-tune LMMs with RL, \emph{Group Relative Policy Optimization} (GRPO)~\cite{shao2024deepseekmathpushinglimitsmathematical} has emerged as a particularly effective method, offering improved training efficiency and greater stability by replacing the PPO value function estimator with group sampling. More specifically, given a class label $c$ and a set of $G$ outputs $\{o_1, o_2, \ldots, o_G\}$ sampled from the policy $\pi_{\theta_{old}}$, the objective of GRPO is to update $\pi_{\theta}$ by maximizing the following function:

\vspace{-10pt}
\begin{equation}
\small
\begin{aligned}
\mathcal{J}(\theta) &= \mathbb{E}_{o_i \sim \pi_{\theta_{old}}} \Bigg[
\frac{1}{G} \sum_{i=1}^{G} \min \bigg(
\frac{\pi_{\theta}(o_i \mid c)}{\pi_{\theta_{old}}(o_i \mid c)},  \\
&\quad \text{clip}\Bigl(
\frac{\pi_{\theta}(o_i \mid c)}{\pi_{\theta_{old}}(o_i \mid c)},
1 - \epsilon, 1 + \epsilon
\Bigr) A_i
\bigg) - \beta D_{\text{KL}} \Bigg],
\end{aligned}
\end{equation}

where $\epsilon$ is the clipping threshold hyperparameter~\citep{schulman2017proximalpolicyoptimizationalgorithms}, $\beta$ controls the strength of the (per token) KL divergence regularization term with respect to the pre-trained policy $\pi_{\theta_{ref}}$ $D_{\text{KL}}(\pi_{\theta} \mid \mid \pi_{\theta_{ref}}) = \frac{\pi_{\theta_{ref}}(o_i \mid c)}{\pi_{\theta}(o_i \mid c)} - \log(\frac{\pi_{\theta_{ref}}(o_i \mid c)}{\pi_{\theta}(o_i \mid c)}) - 1 $, and $A_i$ represents the advantage estimate computed with the rewards obtained within each output group:

\vspace{-10pt}
\begin{equation}
\small
\begin{aligned}
A_i = \frac{r_i-\text{mean}(\{r_1, r_2, ..., r_G\})}{\text{std}(\{r_1, r_2, ..., r_G\})}.
\end{aligned} 
\end{equation}

\subsection{GRPO for Next-Scale Autoregressive Image Models}
\label{subsec:GRPO for Next-Scale Autoregressive Image Models}

Extending GRPO to next-scale autoregressive modelling is straightforward by keeping the Vector Quantized Variational Autoencoder~(VQ-VAE)~\citep{Oord2017NeuralDR} decoder frozen and fine-tuning only the VAR model operating in its discrete latent space. Indeed, in this context, VAR functions exactly like an LLM, except that at each inference step it produces logits for a set of tokens (the next-scale tokens, $r_k$) rather than just one. Therefore, the same GRPO formalization used for LLMs can be applied directly. Specifically, in this context, $\pi_{\theta_{\text{ref}}}$ denotes the pre-trained VAR, while $\pi_{\theta}$ and $\pi_{\theta_{\text{old}}}$ represents the VAR model that we are currently optimizing and sampling from, respectively. The outputs $\{o_1, o_2, \ldots, o_G\}$ correspond to the set of tokens produced autoregressively by $\pi_{\theta_{\text{old}}}$ during sampling, which are subsequently decoded into images by the VQVAE and evaluated by a specific image reward model.

It is important to note that during training, we use simple multinomial sampling from the VAR output logits scaled with a temperature value $\tau$ of 0.7 to balance exploration and exploitation. We deliberately avoid using classifier-free guidance~(CFG)~\citep{ho2022classifierfreediffusionguidance}, top-p or top-k sampling, beam search, or other techniques that could improve sample quality in order to maintain a tractable sampling distribution during optimization. During inference, however, we do use CFG along with top-p or top-k sampling. More concretly, we employ the same hyperparameters as those used in the original VAR model~\citep{VAR}.

Since rewards are typically computed with respect to the final image output, we encounter a credit assignment problem: the individual contribution of each token to the final score is unknown. To address this, we compute the advantages for each sample and apply the same advantage to all tokens. This approximation enables us to maintain a per-token KL penalty, which we found essential for preserving the generation quality of the original model.

\section{Experiments}

For our experiments, we employed the original VAR pre-trained on ImageNet. This model can generate images conditioned on one of the 1,000 original ImageNet labels (it is a class-conditioned image model). We focused on the smallest and largest variants of the model, both of which generate images at a resolution of $256\times256$: VAR-\textit{d}16 (310M parameters) and VAR-\textit{d}30  (2B parameters). 

Unless otherwise explicitly stated, all experiments were conducted by sampling a batch of random labels from the original 1k ImageNet labels and using 8 Nvidia H100 GPUs. Additionally, for all experiments, the following hyperparameters were used: a batch size of 32 labels for sampling, 16 groups, minibatch size of 32 during optimization, a learning rate of $10^{-4}$, $\beta=0.2$, temperature of $0.7$, and $\epsilon=0.2$.

\subsection{Preliminary Experiments: Brightness}

In order to test our implementation, we designed a simple toy experiment where we used GRPO to align the model to generate images with high or low brightness. Brightness is straightforward and inexpensive to compute, and it provides a stable and clear reward signal: $\frac{1}{HW}\sum_{i,j}(0.2989 R_{i,j} + 0.5870 G_{i,j} + 0.1140 B_{i,j})$ where $H=h_K$ and $W=w_K$ denote image height and width, respectively, and $R,G,B$ are the RGB channel values. 

\begin{figure}[thpb!]
  \centering
  \includegraphics[width=0.7\columnwidth]{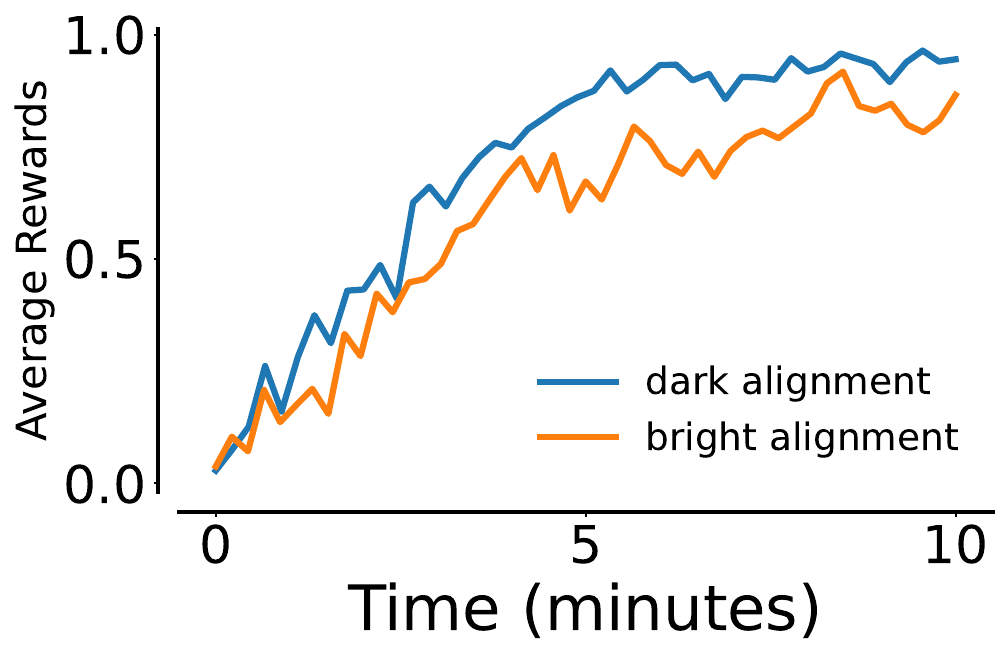}
  \caption{Fine-Tuning VAR-\textit{d}16 with GRPO on dark/bright alignment requires only 10 minutes with a single H100 GPU.}
  \label{fig:brightness-train}
\end{figure}

\begin{figure}[thpb]
  \centering
  \includegraphics[width=\columnwidth]{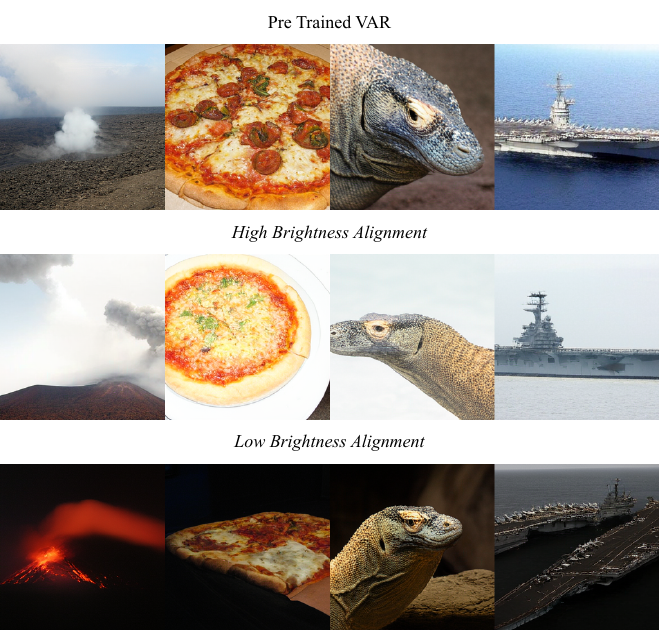}
  \vspace{-10pt}
  \caption{Visual results of the toy experiment aligning the model to generate only bright or dark images.}
  \label{fig:brightness}
  \vspace{-10pt}
\end{figure}

To align the model to generate images with high brightness, we set up a scalar reward giving a value of 1 when brightness is greater than or equal to 0.8, and 0 otherwise. For aligning the model to generate dark images, we give a reward of 1 when brightness is lower than 0.2. As shown in~\autoref{fig:brightness-train}, using GRPO we are able to perfectly align VAR to generate bright or dark images in less than 10 minutes using a single GPU. Despite the simplicity of the task, the method exhibits surprisingly strong sample efficiency and stability, reinforcing our hypothesis that the faster inference speed of VAR makes it well-suited for online RL in image generation. The results of this experiment are shown in \autoref{fig:brightness}.

\subsection{Improving Image Aesthetics}

\begin{figure}[thpb!]
  \centering
  \includegraphics[width=0.7\columnwidth]{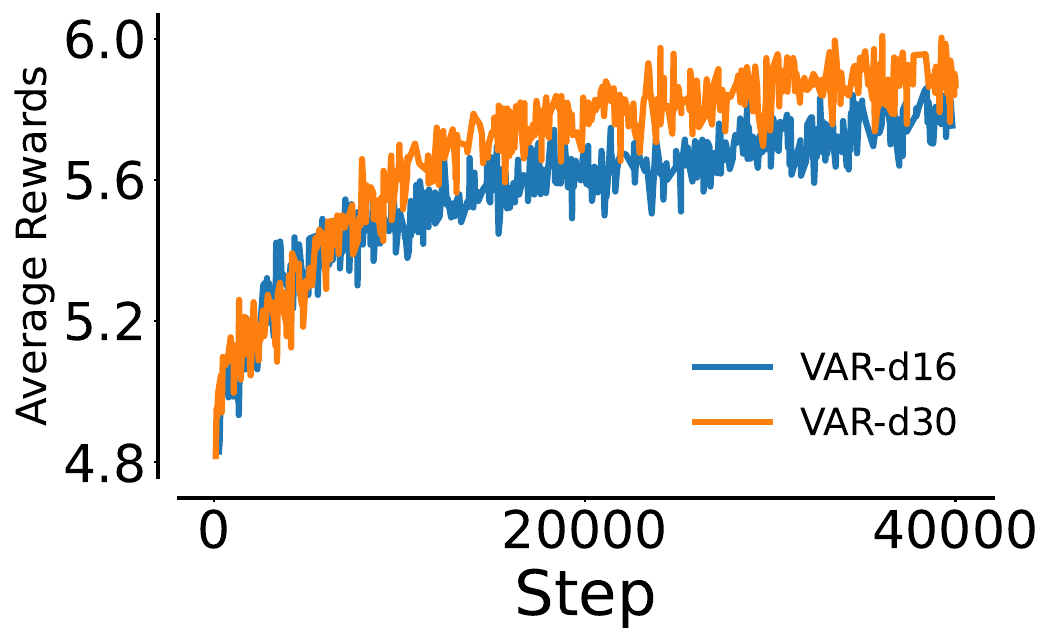}
  \caption{Fine-Tuning VAR-\textit{d}16 and VAR-\textit{d}30 with GRPO using Aesthetic Reward.}
  \label{fig:aes-train}
\end{figure}

\begin{figure}[thpb!]
  \centering
  \includegraphics[width=\columnwidth]{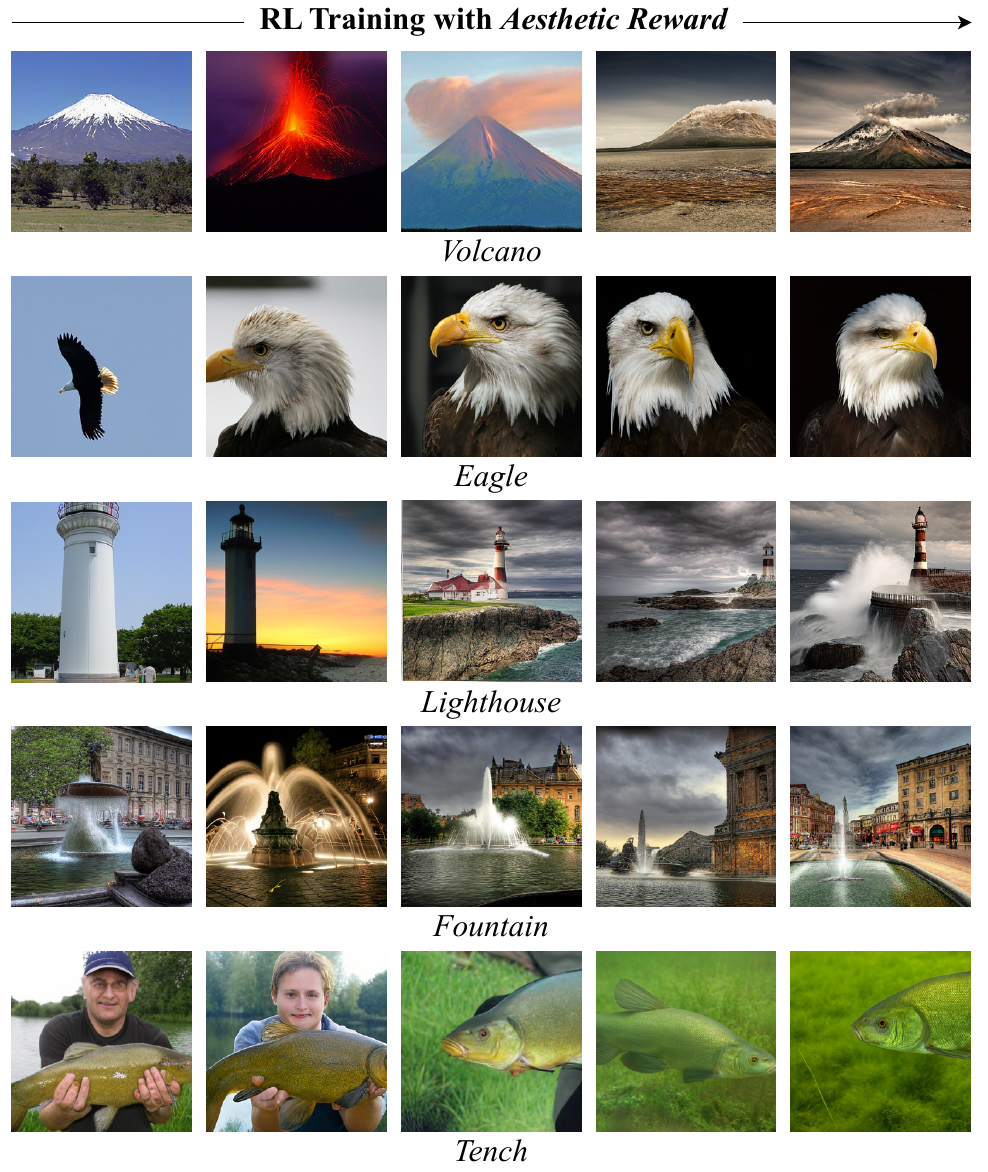}
  \caption{Sampling progress of VAR-\textit{d}30 during GRPO optimization with aesthetic reward.}
  \label{fig:aes-progress}
\end{figure}

To evaluate the capability of GRPO to align the VAR model to more complex scores, we employ LAION’s aesthetic predictor V2~\cite{schuhmann2022laion} (AES) as a proxy for human preferences. AES combines CLIP embeddings with a multi-layer perceptron and was trained on 176,000 images rated by users on an aesthetic scale from 1 to 10, with a score of 10 marking an image as a work of art. For this task, we trained both VAR-\textit{d}16 and VAR-\textit{d}30 for 40,000 steps, which corresponds to a single epoch of VAR's supervised training on ImageNet. This amounted to 16 and 40 hours of training, respectively, using 8 NVIDIA H100 GPUs. As shown in \autoref{fig:aes-train}, both models improve their AES scores by approximately 1 point, with VAR-\textit{d}30 increasing from 4.80 (pre-trained) to 5.80. A progression of the outputs obtained during the optimization process is shown in \autoref{fig:aes-progress}.

In order to validate that the fine-tuned models are still capable of generating images corresponding to the ImageNet labels, we can no longer rely on the FID score, as the RL optimization has altered the generative distribution of ImageNet. Instead, we employ a standard ImageNet classifier (ResNet50 IMAGENET1K\_V2~\cite{vryniotis2021train}) and observe the difference in accuracy between samples from the reference and fine-tuned models. In \autoref{tab:aes-table}, we report the Aesthetic Score computed from 10,000 samples (10 per class) for the pre-trained and fine-tuned models, along with the ResNet50 top-5 accuracy. Notably, the models fine-tuned with GRPO maintain an accuracy above 90\% while generating images with an Aesthetic Score approximately 1 point higher than that of the ImageNet validation set. This observation confirms that the model did not diverge significantly from the ImageNet distribution during training.

In \autoref{tab:aes-table}, we also report the scores for VAR-\textit{d}16 fine-tuned using only 50\% of the ImageNet labels. In this case, the AES score increased more moderately by 0.5 points. However, it is noteworthy that the AES score also increased by 0.5 points for the labels not seen during training, suggesting that the model generalized effectively during the fine-tuning process.

\begin{table}[hbtp!]
\caption{Comparison of models by Aesthetic Score and ResNet50 Accuracy@5. Notice that the very high accuracy observed for the pre-trained model reflects its ability to reproduce the distribution of the ImageNet training set, where ResNet50 IMAGENET1K\_V2 likely achieves near-perfect accuracy. VAR-\textit{d}16 GRPO (seen 50\% labels) was fine-tuned using 50\% of the ImageNet classes, while the other GRPO models used all possible labels.}
    \centering
    \resizebox{\columnwidth}{!}{%
    \begin{tabular}{|l|c|c|}
        \hline
        \textbf{Model} & \textbf{Aesthetic Score} & \textbf{ResNet50 Accuracy@5} \\
        \hline
        ImageNet Validation Set & 4.91 & 95.43\% \\
        \hline
        VAR-\textit{d}16~\citep{VAR} & 4.64 & 97.8\% \\
        VAR-\textit{d}16 GRPO (seen 50\% labels) & 5.10 & 97.7\% \\
        VAR-\textit{d}16 GRPO (unseen 50\% labels) & 5.09 & 97.9\% \\
        VAR-\textit{d}16 GRPO (seen 100\% labels) & \textbf{5.55} & 95.92\% \\
        \hline
        VAR-\textit{d}30~\citep{VAR} & 4.80 & 99.8\% \\
        VAR-\textit{d}30 GRPO (seen 100\% labels) & \textbf{5.80} & 90\% \\
        \hline
    \end{tabular}
    }
    \label{tab:aes-table}
\end{table}

\subsection{Learning to Paint with CLIP Alignment}

To further evaluate the capabilities of our method, we employ the CLIP Score~\cite{radford2021learning} to measure the similarity between the generated samples and a fixed text prompt. In this experiment, our goal is to assess the ability of GRPO to align a generative model with respect to a different modality (text), even though this modality does not directly condition the generative process. Specifically, the VAR model remains conditioned solely on the input labels, while the CLIP Score is computed relative to a fixed text prompt.

\begin{figure}[thpb]
  \centering
  \includegraphics[width=0.7\columnwidth]{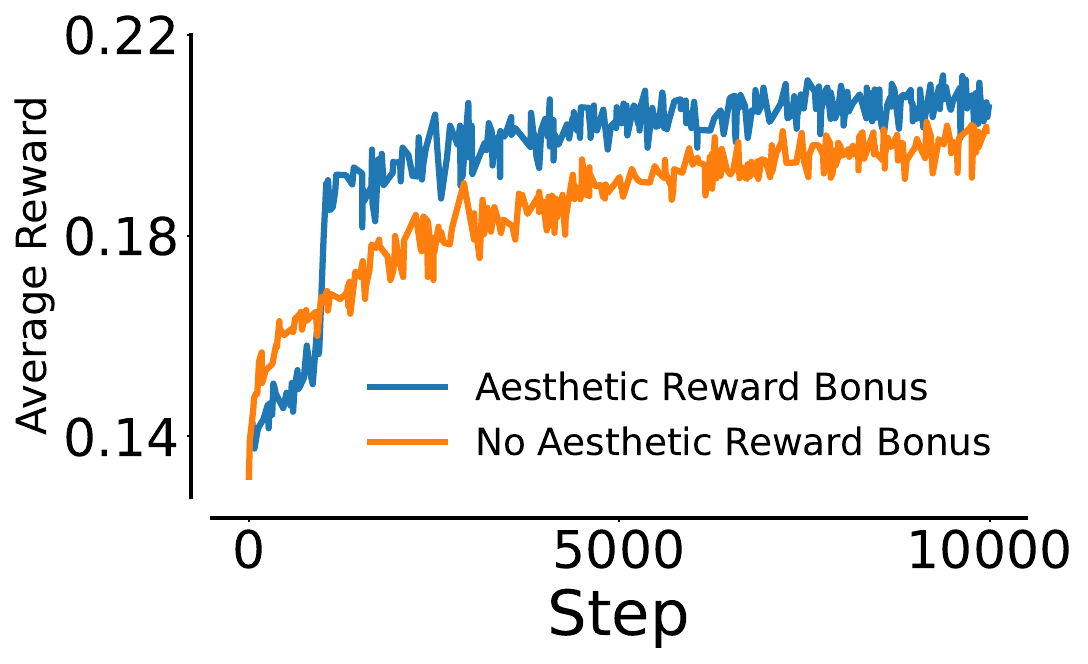}
  \caption{CLIP Score measuring alignment with the prompt “A painting” during fine-tuning, with and without the application of the Aesthetic Reward bonus (excluded from the plots).}
  \label{fig:clip-painting-rewards}
\end{figure}

We begin by conditioning the CLIP score on the prompt \textit{“an old picture”}. Since the ImageNet training set contains images of this type, the model quickly generates samples that align with the style. We therefore move to a more challenging prompt: \textit{“a painting”}. Note that the original ImageNet dataset discouraged paintings or drawings, since annotators were instructed to collect: ``photos only, no painting, no drawings, etc.''~\cite{deng2012large}. This means that, in order to align with this kind of prompt, the model should in principle generate samples that are out of the training distribution. Nevertheless, as shown in \autoref{fig:clip-painting-rewards} after some iterations, the policy finds ways to increase the ``painting'' CLIP Score. This suggests that, at least in part, the model really learned something new through RL exploration and exploitation, i.e., \textit{it learned artistic patterns without demonstrations}. Obviously, this kind of hypothesis should be explored more deeply, but the preliminary results obtained in this context seem to indicate that online RL could, in principle, align an image generative model even further than what has been seen in the training data.

\begin{figure}[thpb]
  \centering
  \includegraphics[width=\columnwidth]{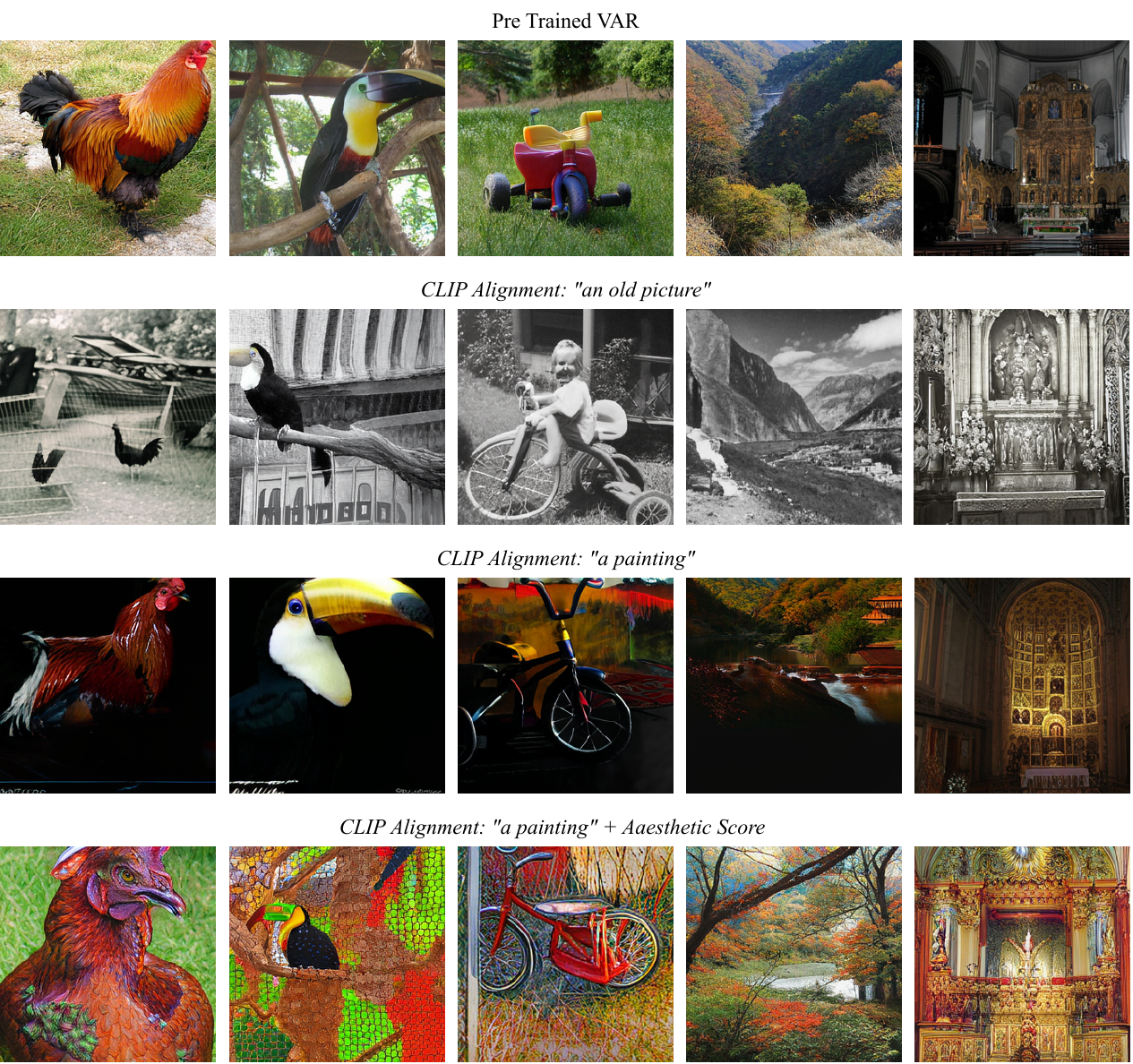}
  \vspace{-10pt}
  \caption{Experiments conducted to align VAR with a fixed prompt using the CLIP Score. In this setup, VAR itself is not conditioned on the prompt; only the reward model is. Since VAR was trained on ImageNet, the training data predominantly consists of photographs (no paintings, no drawings). This means that the artistic effects observed in the samples generated when aligning the model with the caption \textit{“a painting”} should be attributed primarily to RL-driven exploration and optimization, as if the model effectively “learned to paint” during fine-tuning. Notice how the model adopts a simple yet effective approach to increase the “picture” reward by incorporating a technique reminiscent of Caravaggio’s chiaroscuro. This effect is suppressed when an Aesthetic Score is added to the reward, resulting in more colorful and stylistically diverse images.
}
  \label{fig:clip-alignment}
\end{figure}

We display the results in ~\autoref{fig:clip-alignment}. These experiments were performed with VAR-\textit{d}30 fine-tuned for 10k steps, corresponding to around 10 hours of compute in 8 H100 GPUs.

\subsection{Ablations}

Finally, we provide a short discussion on hyperparameter ablations which may be useful for future researchers.

\begin{figure}[thpb]
  \centering
  \includegraphics[width=\columnwidth]{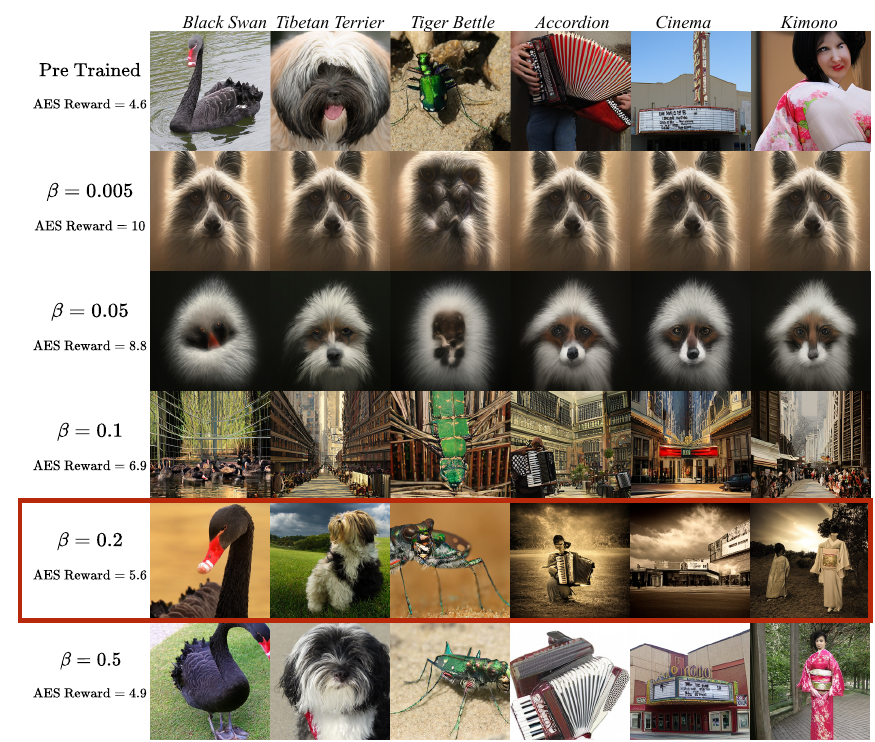}
  \caption{Effect of $\beta$ KL coefficient when training with GRPO.}
  \vspace{-10pt}
  \label{fig:aes-kl}
\end{figure}

\paragraph{KL Penalty} The importance of the KL penalty in GRPO for image generation is illustrated in \autoref{fig:aes-kl}. Without sufficient regularization, the model easily finds a reward-hacking policy and loses the ability to generate images properly conditioned on the labels. The KL term also acts as an entropy bonus, discouraging the model from generating identical or highly similar images even for different labels.

\begin{figure}[thpb]
  \centering
  \includegraphics[width=0.7\columnwidth]{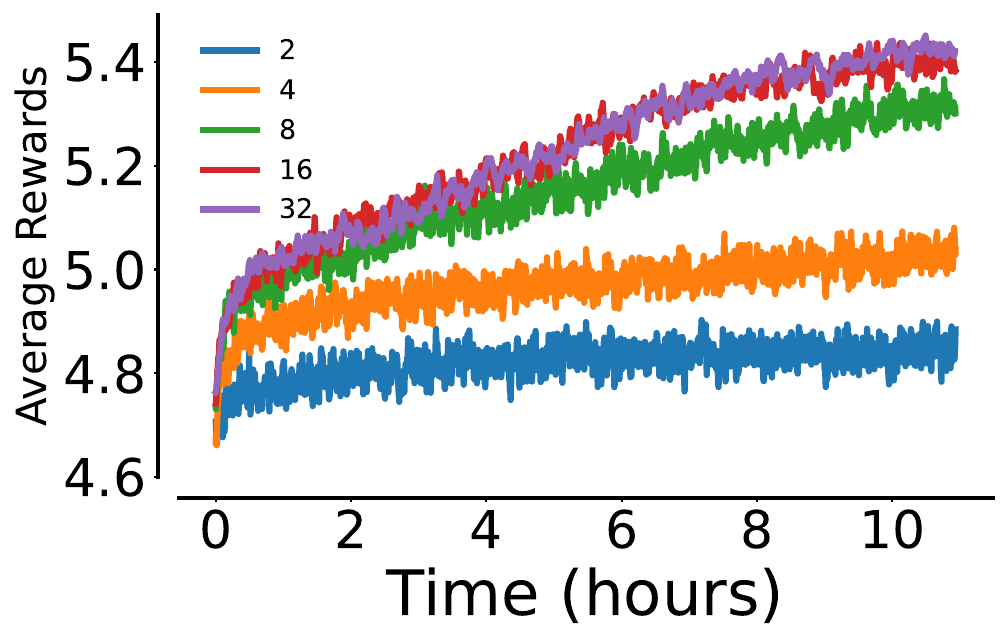}
  \caption{Effect of varying the number of groups during the sampling phase of GRPO (Aesthetic Reward optimization).}
  \label{fig:aes-numgroups}
\end{figure}

\paragraph{Group Optimization} In \autoref{fig:aes-numgroups}, we demonstrate how group optimization effectively aids training. Fixing a compute time budget, we find that increasing the number of groups during sampling monotonically enhances the performance of GRPO, at least up to 16 groups.

\section{Conclusion}

In this work, we have demonstrated the effectiveness and efficiency of GRPO for aligning next-scale visual autoregressive models. A natural next step is to extend this method to text-to-image VAR-based models~\cite{Infinity, voronov2024switti}.

\small
\bibliography{example_paper}
\bibliographystyle{icml2025}

\newpage
\appendix
\onecolumn


\end{document}